%% file: hicss51.tex
\documentclass[10pt]{article}
\input{hicss51-packages.tex}

\usepackage{rotating}
\usepackage{wasysym}
\usepackage{array}
\newcommand*{\Scale}[2][4]{\scalebox{#1}{$#2$}}%
\newcolumntype{P}[1]{>{\centering\arraybackslash}p{#1}}

\title{A New Metric for Lumpy and Intermittent Demand Forecasts: Stock-keeping-oriented Prediction Error Costs}

\author{Dominik Martin \\
  Karlsruhe Institute of Technology \\
  Kaiserstr 89, 76133 Karlsruhe \\
  Germany \\
  {\underline{martin@kit.edu}} \\\And
  Philipp Spitzer \\
  Karlsruhe Institute of Technology \\
  Kaiserstr 89, 76133 Karlsruhe \\
  Germany \\
  {\underline{office@ksri.kit.edu} }\\\And 
  Niklas K\"uhl \\
  Karlsruhe Institute of Technology \\
  Kaiserstr 89, 76133 Karlsruhe \\
  Germany \\
  {\underline{kuehl@kit.edu}} \\}

\date{}

\begin{document}
\maketitle
\begin{abstract}
Forecasts of product demand are essential for short- and long-term optimization of logistics and production. Thus, the most accurate prediction possible is desirable. In order to optimally train predictive models, the deviation of the forecast compared to the actual demand needs to be assessed by a proper metric. However, if a metric does not represent the actual prediction error, predictive models are insufficiently optimized and, consequently, will yield inaccurate predictions. The most common metrics such as MAPE or RMSE, however, are not suitable for the evaluation of forecasting errors, especially for lumpy and intermittent demand patterns, as they do not sufficiently account for, e.g., temporal shifts (prediction before or after actual demand) or cost-related aspects.
Therefore, we propose a novel metric that, in addition to statistical considerations, also addresses business aspects. Additionally, we evaluate the metric based on simulated and real demand time series from the automotive aftermarket.

\end{abstract}

\section{Introduction}
Forecasts are crucial in making accurate decisions about the future when it is uncertain. Accurate forecasting is considered an important prerequisite for effective planning and organization in various areas such as finance, supply chain management, sales, meteorology and many others.

In order to meet the manifold characteristics of various domain-specific forecasting problems, there are several techniques available for estimating future conditions. Depending on which forecasting method is chosen one would evaluate which method appears to be most appropriate for the underlying problem formulation. Hence, to estimate the performance of forecasting methods and to evaluate predictions accordingly, various accuracy measures have been proposed and intensively discussed over the last couple of decades. However, there is no overall best single performance metric which can be applied universally to any kind of forecast problem \cite{Mahmoud1984}. 

Depending on the chosen metric, forecasts can yield completely different performances, which makes an appropriate evaluation extremely difficult. Additionally, some metrics are not suitable for the application on certain data. For example, the mean absolute percentage error (MAPE) produces infinite or undefined values when actual values are zero or close to zero which often is the case in certain applications \cite{Kim2016}. Hence, most existing metrics work only with smooth and linear patterns, however, become less precise or even unusable with an increasing frequency of intermittent patterns \cite{Wallstrom2010}. Product demand forecasting in particular is one such case of application since products with intermittent demand tend to have many occurrences of zero values \cite{Waller2016}. Evaluating forecasts of stock-keeping units with traditional accuracy metrics can result in misleading findings which creates the need to develop appropriate error measures in such industry fields by including further factors. \cite{Kourentzes2013}.

Intermittent demand is characterized by non-demand intervals, whereas lumpy demand is characterized by large fluctuations in the size of actual demand occurrences (see Figure~\ref{fig: lumpy}) \cite{Wallstrom2010, Mukhopadhyay2012}.

\begin{figure}[thb]
	% Use the relevant command to insert your figure file.
	% For example, with the graphicx package use
    \centering
	\includegraphics[clip,width=0.93\linewidth]{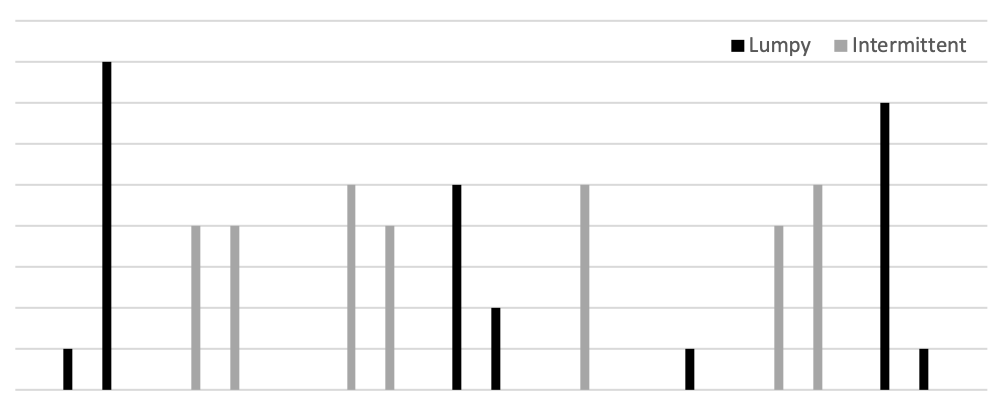}
	% figure caption is below the figure
	\caption{Exemplary lumpy and intermittent patterns}
	\label{fig: lumpy}       % Give a unique label
\end{figure}

In this work, we present various commonly applied metrics for the evaluation of forecast performance and discuss the advantages and disadvantages of each. In addition, following the Design Science Research paradigm according to Vaishnavi and Kuechler \cite{Vaishnavi2004}, we propose a new metric that, in addition to statistical considerations, also addresses business aspects for this particular type of sporadic demand pattern. In comparison to other proposed measurements this metric takes into account horizontal and vertical shifts in prediction over the forecasting horizon. This means, not only errors of specific points in time are considered but also temporal interaction between predictions of different points in time. Our proposed metric measures the state of outstanding orders or units in stock by considering opportunity and stock-keeping costs. Therefore, this metric depicts the actual forecasting error in terms of business and cost factors and constitutes an appropriate decision support tool.

The remainder of this work is organized as follows. In Section \ref{sec: related work}, we explore existing literature about performance metrics for intermittent and lumpy demand forecasting. Section \ref{sec: suggestion} addresses the need for a suitable performance measure by proposing the \emph{Stock-keeping-oriented Prediction Error Costs (SPEC)} metric. In Section \ref{sec: evaluation}, the SPEC metric is evaluated by means of synthetic data covering various possible cases and real demand data from the automotive after-market. Section 5 provides a robustness check on a different data set. Finally, we form a conclusion in Section \ref{sec: conclusion}.

\section{Related Work and Research Gap}
\label{sec: related work}
In literature there are many different metrics used to evaluate forecasting accuracies. Assessment of forecasts depends on the choice of a proper metric. In different studies a variety of accuracy measures have been applied \cite{yan2012toward, makridakis1982accuracy, WERON2014, HONG2014357, kalekar2004time}. Table~\ref{tab: metrics} compares the most common representatives and their characteristics. Full circles represent a satisfied criterion, while empty circles imply that a characteristic is not met. Half circles, on the other hand, mean that no clear statement is possible. 

Simultaneous simple yet unreliable absolute error measures are used particularly frequently. They are based on the absolute error between the actual value at a time and its associated prediction. 
Different authors investigate the capabilities of such measures to evaluate forecasting techniques and compare them with another. The authors Chen and Yang \cite{Chen2004} claim that the root mean squared error (RMSE), for instance, is not an appropriate metric to compare different forecasting methods since the metric is not unit-free. Armstrong and Collopy \cite{Armstrong1992} state the high sensitivity against outliers. The disadvantages of these metrics being scale-dependent and highly sensitive against outliers are addressed by percentage errors \cite{Shcherbakov2013}. 

Percentage errors have the advantage of being scale-independent because the forecast error is divided by the actual value \cite{Hyndman2006}. However, this creates the disadvantage that these metrics are infinite or undefined if the actual values are zero or close to zero \cite{Shcherbakov2013}. Especially in scenarios where demands with actual values likely being zero should be predicted, representatives like MAPE or RMSPE (root mean squared percentage error) are not proper choices for evaluation purposes \cite{Armstrong1992, Hyndman2006}. Additionally, percentage errors have the disadvantage of putting a heavier penalty on positive errors than on negative errors which leads to asymmetry. In order to overcome this issue, so-called symmetric error measures have been proposed \cite{MAKRIDAKIS1993}. Despite its name the symmetric mean absolute percentage error (sMAPE) is also assymetrical and prone to infinity \cite{Goodwin1999}. 

\begin{table*}[thb]% 
    \centering % Center table
\begin{tabular}{p{3cm}|P{1.2cm}P{1.2cm}P{1.2cm}P{1.2cm}|P{1.2cm}P{1.2cm}}
\multicolumn{1}{c|}{} & \multicolumn{4}{c|}{statistical aspects} & \multicolumn{2}{c}{business aspects}\\
  & \rotatebox{90}{ scale independency } & \rotatebox{90}{ no division by zero } & \rotatebox{90}{ outlier insensitivity } & \rotatebox{90}{ symmetry } & \rotatebox{90}{ interpretability } & \rotatebox{90}{ economical considerations } \\ \hline
Absolute Errors & & & & & & \\ 
 - MAE / MdAE & \Circle & \CIRCLE & \Circle & \CIRCLE & \CIRCLE & \Circle \\ 
 - MSE & \Circle & \CIRCLE & \Circle & \CIRCLE & \LEFTcircle & \Circle \\ 
 - RMSE & \Circle & \CIRCLE & \Circle & \CIRCLE & \LEFTcircle & \Circle \\ 
  Percentage Errors & & & & & & \\ 
 - MAPE / MdAPE & \CIRCLE & \Circle & \Circle & \Circle & \CIRCLE & \Circle \\ 
 - RMSPE & \CIRCLE & \Circle & \Circle & \Circle & \LEFTcircle & \Circle \\ 
  Symmetric Errors & & & & & & \\ 
 - sMAPE & \CIRCLE & \Circle & \Circle & \Circle & \Circle & \Circle \\
  Scaled Errors & & & & & & \\ 
 - MASE & \Circle & \Circle & \CIRCLE & \CIRCLE & \LEFTcircle & \Circle \\ 
 - RMSSE & \Circle & \Circle & \CIRCLE & \CIRCLE & \LEFTcircle & \Circle \\ \hline 
 SPEC (this work) & \CIRCLE & \CIRCLE & \LEFTcircle & \CIRCLE & \CIRCLE & \CIRCLE       
\end{tabular}
    \caption{Properties of traditional metrics and SPEC}
    \label{tab: metrics} 
\end{table*}

Scaled measures try to overcome the scaling issues by comparing the forecasts with another benchmark forecast method; usually the na\"{\i}ve one \cite{Hyndman2006}. Additionally, metrics like mean absolute scaled error (MASE) are symmetric and insensitive to outliers. Nevertheless, measures such as MASE or RMSSE (root mean squared scaled error) can be undefined because they are divided by zero \cite{Shcherbakov2013}. 

In terms of interpretability, metrics based on unscaled absolute errors like MAE or MAPE are well comprehensible. Contrastingly, scaled and squared measures (MSE, RMSE, RMSPE, MASE, RMSSE) are relatively difficult to interpret because they generate distorted values \cite{Shcherbakov2013, Hyndman2006}.

In addition to the metrics discussed in this section, there are a number of other metrics that try to address disadvantages of others by, for example, choosing the median as an aggregation technique instead of the mean. The resulting metrics, such as the median absolute percentage error (MdAPE), however, can hardly be distinguished from the original variants in their characteristics \cite{Shcherbakov2013, Hyndman2006}.

Especially in the case of lumpy or intermittent demand time series which are characterized by many zero values, the division by zero is particularly critical. Another disadvantage arises if classic statistical metrics are used. Most metrics are based on isolated deviations between prediction and actual value without taking the dependency between historical and future values into account. This is due to calculating the mean over these deviations throughout every single time step in the time series. Wallstr\"om and Segerstedt \cite{Wallstrom2010} present a metric that avoids such limitations. In their work on supply chain prediction systems the so called \emph{periods in stock} measure counts the number of periods that goods spend in a fictive warehouse and the number of periods, goods are not available due to absence in stock. According to the authors, big values emerge due to overestimating demand as well as big negative values from underestimating demand. However, a limitation of their work is that over- and underestimation of demand can also result in small values of the metric in which the result does not represent the actual quality of the forecast leading to a possibly biased evaluation criterion. 

We propose integrating the economic aspects of this metric into a novel metric that avoids the shortcomings of the statistical nature of periods in stock as well as of traditional metrics.

\section{Suggestion and Development}
\label{sec: suggestion}

In order to address the lack of applicability of existing metrics on demand time series with the special characteristic of frequent zero values, we propose a new metric, which accounts the shortcomings stated in the previous section as well as economic aspects in particular.

The purpose of this new metric is to measure the prediction accuracy by comparing actual demand and forecast in the form of theoretically incurred costs over the forecast horizon. In retail, stock-keeping costs are incurred for stock in the warehouse. The greater the amount and longer the duration the inventory stays in the warehouse, the higher the stock-keeping costs. In addition, opportunity costs arise in the form of unsatisfied customer orders if requested items are not available in the warehouse. The proposed metric should be zero for optimal storage by a perfect prediction and if the deviation from the perfect prediction increases, the metric should be greater than zero, representing the costs of the misprediction.

The calculation of the metric is based on the assumption that the stock is depicted by forecasts and demands in a fictitious stock. Accordingly, forecasts represent deliveries, and thus storage in
the warehouse at a given time. Demands, however, represent departures from the warehouse for the respective time. If, according to this logic, an item is unavailable over several units of time in the warehouse, the forecast should be penalized more severely. Accordingly, unsatisfied orders arising from zero stock of an item will also be penalized more heavily the longer the order can not be fulfilled.

Thus, our proposed metric measures a cost equivalent caused by a forecasting error, which is composed of stock-keeping and opportunity costs. Because of this special characteristic we introduce the \emph{Stock-keeping-oriented Prediction Error Costs (SPEC)} metric which is calculated by the equation in Equation~\ref{equ: spec_equation}.

\begin{figure*}[thb]
\renewcommand{\figurename}{Equation}%
\setcounter{figure}{0}
    \centering
	\[
	\Scale[0.985]{
	\displaystyle
    SPEC_{\alpha_1,\alpha_2} = \frac{1}{n}
     \sum_{t=1}^{n} \sum_{i=1}^{t}  \left ( max \left [ 0; min \left [ y_i ; \sum_{k=1}^{i} y_k - \sum_{j=1}^{t} f_j  \right ]  \cdot  \alpha_1 ;  min \left [ f_i ; \sum_{k=1}^{i} f_k - \sum_{j=1}^{t} y_j  \right ]  \cdot  \alpha_2  \right ] \cdot (t-i+1) \right )
     }
    \]
    \caption{Stock-keeping-oriented Prediction Error Costs (SPEC)}
	\label{equ: spec_equation}       % Give a unique label
\end{figure*} 

The length of the time series is labeled by $n$, whereas the actual demand at time $t$ is characterized by $y_t$ and the corresponding forecast by $f_t$. The opportunity and stock-keeping costs are defined by the parameters $\alpha_1 \in [0, \infty]$ and $\alpha_2 \in [0, \infty]$, respectively. Our recommendation is to choose $\alpha_1$ and $\alpha_2$ so that their sum is 1. Thus, we recommend the relationship $\alpha_1 = 1-\alpha_2$ to ensure comparability despite a possible changing cost ratio. However, any values greater than zero can be chosen to reflect business-specific monetary cost equivalents.

SPEC calculates an error term for each time step of the forecast. For each of these time steps either opportunity costs or stock-keeping costs arise, but never both at the same time. The inner max-min-term of the equation specifies if either of them exists at a given time step. Iterating over each time step beginning with the first element, the equation penalizes every fictive Stock Keeping Unit (SKU) that is yielding to this gap in stock at the current time step. By doing so, each time step is assessed so the actual demand is forecasted accurately up to that point in time. Underestimation is accounted for the left minimum function counting SKUs not available in a fictive stock and therefore creating opportunity costs. In case of overestimation of the actual time series, the right minimum function counts all SKUs sitting in that fictive stock. Such a characteristic resembles situations where the forecast is penalized with stock-keeping costs.  With the aid of this approach, the metric is able to account the forecast of the entire forecasting horizon instead of inspecting on specific time steps only. Nevertheless, SPEC may also penalize time steps at which actual value and forecast coincide. However, due to inaccurate prediction in the entire time frame the seemingly perfect forecast at this time step results in costs, as SKUs are already sitting in stock or are unavailable in situations of coming demand. Hence, the evaluation of the forecast is put into the context of the time series as a whole. Especially in situations where not only one-step ahead forecasts are relevant but n-step ahead forecasts are needed, SPEC overcomes issues of traditional error metrics applied in past studies.   

While traditional metrics only address forecast errors in isolation, the SPEC metric also includes the deviation of the prediction time (prediction error in x-direction) in addition to the vertical deviation (prediction error in y-direction).

By utilizing SPEC it is possible to measure forecasts two-dimensionally as the magnitude and time difference are taken into account. In scenarios where predictions are based on SKUs a pure focus on the difference of actual value and forecast may be insufficient for a proper evaluation of such a time series. When demand is not predicted accurately, costs not only arise at this particular time step but comprise  stock-related costs over time. SPEC is overcoming these shortcomings by assessing both, x- and y-direction of time series forecasts.

\begin{figure*}[thb]
	% Use the relevant command to insert your figure file.
	% For example, with the graphicx package use
    \centering
	\includegraphics[clip,width=0.93\linewidth]{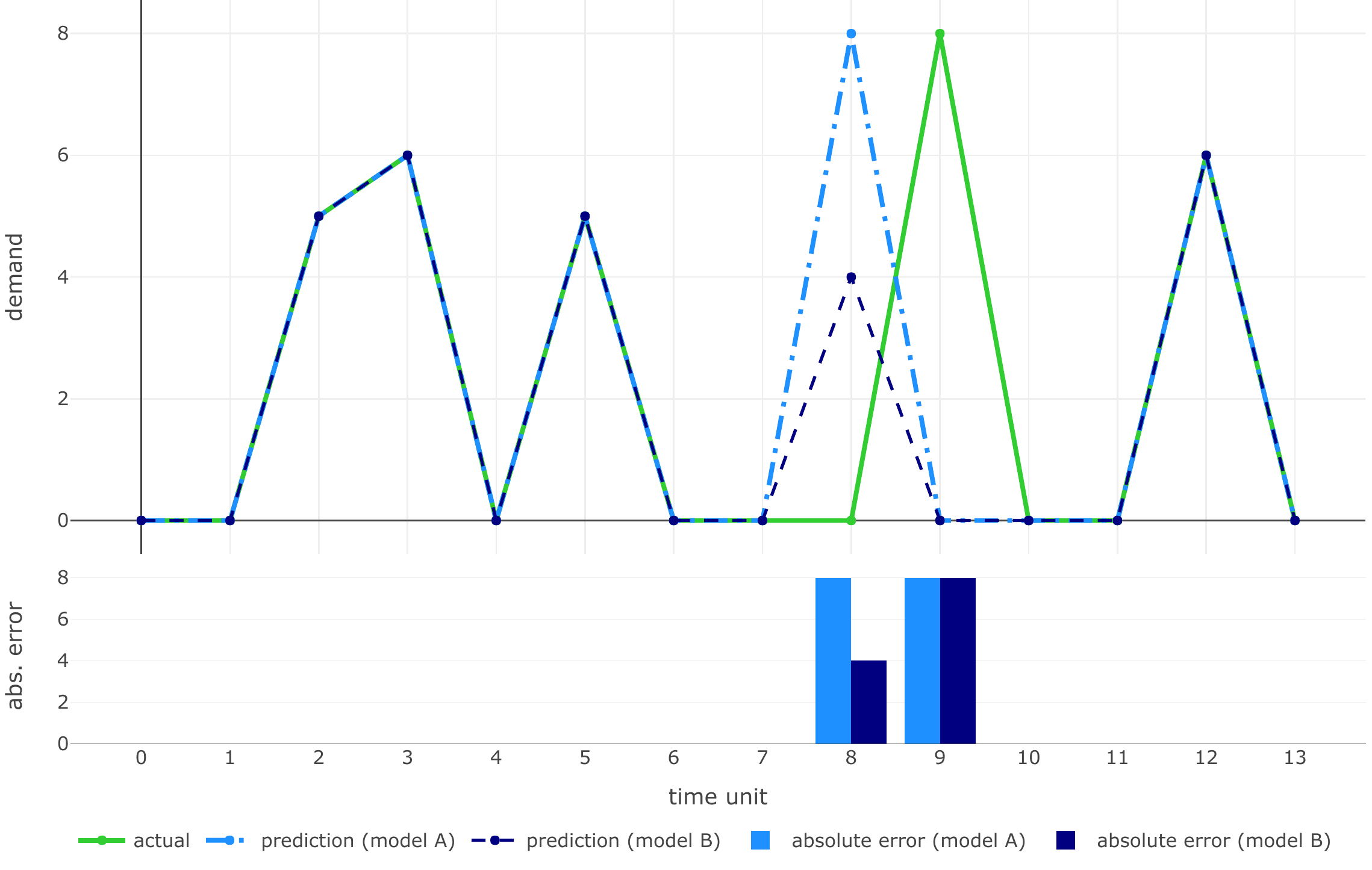}
	% figure caption is below the figure
	\caption{Exemplary demand time series and associated predictions (top) as well as absolute errors (bottom) of two exemplary predictive models}
	\label{fig: spec_sample}       % Give a unique label
\end{figure*}

Figure~\ref{fig: spec_sample} on page \pageref{fig: spec_sample} illustrates this issue with an exemplary time series (green) and two associated predictions (light blue and dark blue) of two fictive models. In both examples, the actual demand is forecasted almost perfectly, however, the demand (8 units) at time unit 9 is predicted one time unit too soon by model~A and additionally 4 units too few by model~B. Evaluating the prediction quality of model~A particularly, this means that storage costs would be incurred for 8 units for exactly one time unit.

While the RMSE for the prediction of model~A depicted in Figure~\ref{fig: spec_sample} is 3.02 (see Table~\ref{tab: comparison}), the $SPEC_{0.75,0.25}$ rates the error at 0.14. Would the misprediction not be shifted to the left by one time unit, but be overestimated by 11 units at time unit 9 instead, the RMSE results in an almost equal value of 2.94. However, the $SPEC_{0.75,0.25}$ of 2.30 represents the clear overestimation of demand  compared to the actual demand more accurately. Another factor arising when incorporating the forecast of model B depicts the property of comparing different models with another. Applying error metrics can be utilized not only for assessing one model's performance but also for benchmarking against other models. The goal of such comparisons constitutes the determination of the model which best fits to make predictions for a given time series. Considering the bottom part of Figure~\ref{fig: spec_sample} it becomes clear that model~B causes an overall smaller absolute error than model~A, which is also reflected in smaller values of the traditional error measures (see e.g., MAE, RMSE, MASE in Table~\ref{tab: comparison}). However, it is obvious that model A better estimates the actual demand, as the lack of demand is compensated after only one unit of time, while model~B still lacks four units. Thus, in this example model~B achieves a $SPEC_{0.75,0.25}$ of 2.00 (worse than for model~A) while the RMSE is 2.39 (better than model~A). This is due to the fact that the forecast of model~B clearly underestimates the actual demand since four SKUs are not available and corresponding orders cannot be fulfilled until the parts arrive in the warehouse. While the RMSE actually states the outcome of model~B as the more accurate one, $SPEC_{0.75,0.25}$ properly assigns model~A as a more precise prediction tool when considering the actual time series. Thus, this example shows the evaluation property of $SPEC_{0.75,0.25}$ when assessing several models in situations with incoming warehousing costs. Our metric better addresses this than other traditional metrics, like RMSE.

\begin{table}[thb]% 
    \centering % Center table
\begin{tabular}{lrr}
\hline
      & Model A & Model B \\ \hline
MAE   & 1.143   & 0.857   \\
RMSE  & 3.024   & 2.390   \\
MAPE  & inf     & inf     \\
sMAPE & 0.667   & 0.667   \\
MASE  & 0.297   & 0.223   \\
$SPEC_{0.75,0.25}$  & 0.143   & 2.000   \\ \hline
\end{tabular}
\caption{Results of different metrics scoring forecast errors of exemplary models depict in Figure~\ref{fig: spec_sample}}
\label{tab: comparison} 
\end{table}

\begin{figure*}[thb]
	% Use the relevant command to insert your figure file.
	% For example, with the graphicx package use
    \centering
	\includegraphics[clip,width=0.93\linewidth]{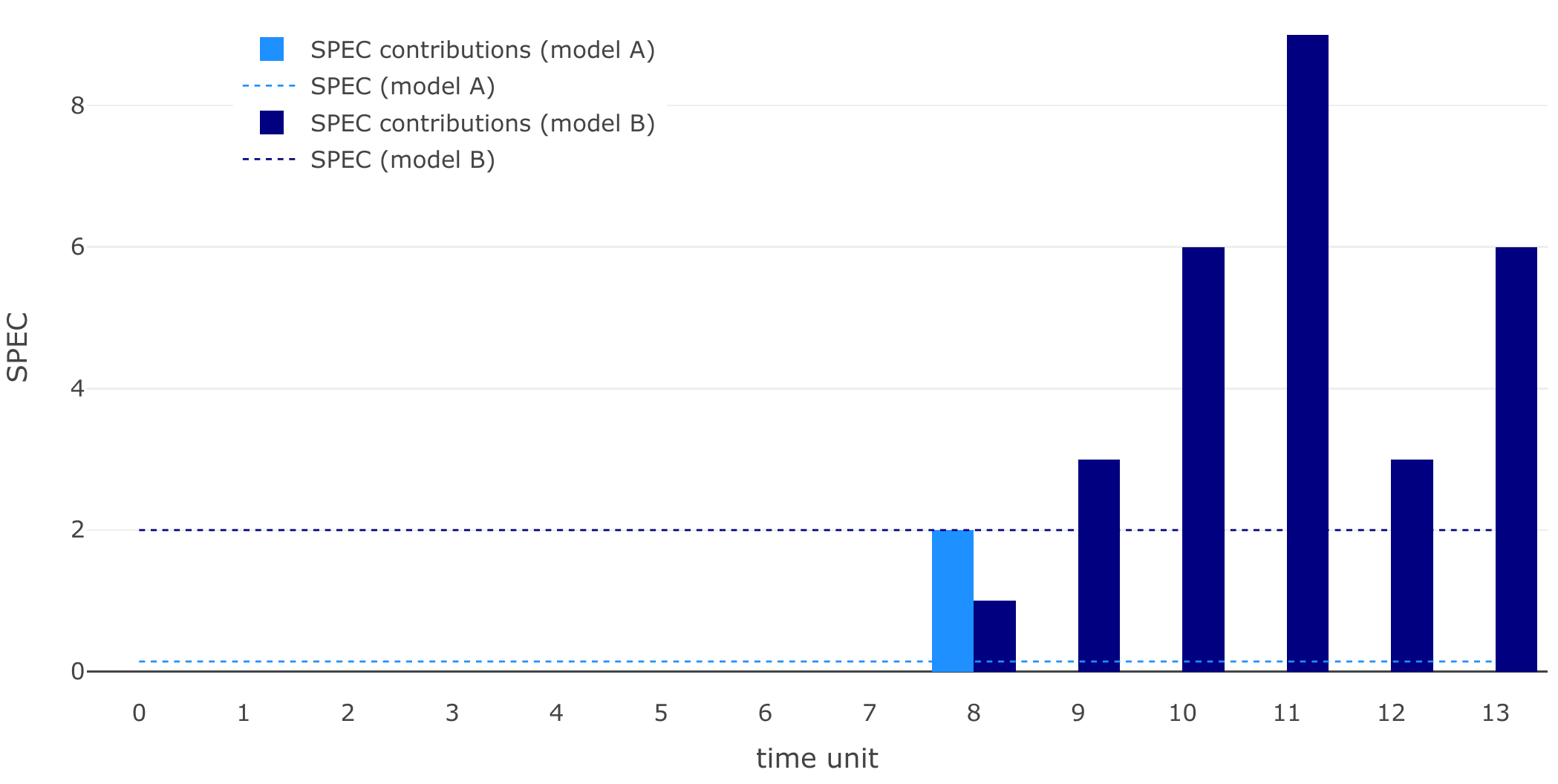}
	% figure caption is below the figure
	\caption{Composition of exemplary SPEC values according to exemplary models depict in Figure \ref{fig: spec_sample}}
	\label{fig: spec_comp}       % Give a unique label
\end{figure*} 

Figure~\ref{fig: spec_comp} depicts the composition of the SPEC values of the models illustrated in Figure~\ref{fig: spec_sample} by their individual contributions per time unit. 
Considering the evolvement of SPEC over the forecasting horizon as shown in Figure~\ref{fig: spec_sample} for both models we can see that at time unit 8 the prediction of model~A leads to a stock level of eight units. These eight units are kept in stock until demand occurs at time unit 9. Overall, for model~A we receive stock-keeping costs of 2 ($=\alpha_2 \cdot 8=0.25 \cdot 8$) at time unit 8 which lead to a $SPEC_{0.75,0.25}$ of 0.14 (mean over complete forecast horizon). Since model~B only predicts half the amount at time unit 8 compared to model~A the stock-keeping costs for time unit 8 result in a value of 1. Nevertheless, demand at time unit 9 cannot be fulfilled completely since too few units are in stock. This leads to opportunity costs of 3 ($=\alpha_1 \cdot 4=0.75 \cdot 4$), since the weighing factor addressing the opportunity costs $\alpha_1$ is 0.75. Furthermore, at time unit 11 there are four units for three days missing which is why opportunity costs of 9 occur. First at time unit 12 model~B predicts six units but at the same time six units are inquired. Therefore the four units, which were requested at time unit 9 are fulfilled first. Simultaneously, two units of the new occurring demand can be met additionally. Hence, a demand of four cannot be satisfied leading to opportunity costs of 3 at time unit 12. Comparing the SPEC approach to other accuracy measures like the absolute error, the latter is accounting the prediction shift at time unit 8 and 9 of model~A twice which leads to an incorrect evaluation of the forecasts. Since model~B only predicts half the amount at time unit 8 compared to model~A the absolute error diminishes. However, in terms of stock-keeping units and upcoming demand the prediction of model~A leads to a more precise result compared to model~B since orders can be fulfilled faster.

\begin{figure*}[thb]
	% Use the relevant command to insert your figure file.
	% For example, with the graphicx package use
    \centering
	\includegraphics[clip,width=0.93\linewidth]{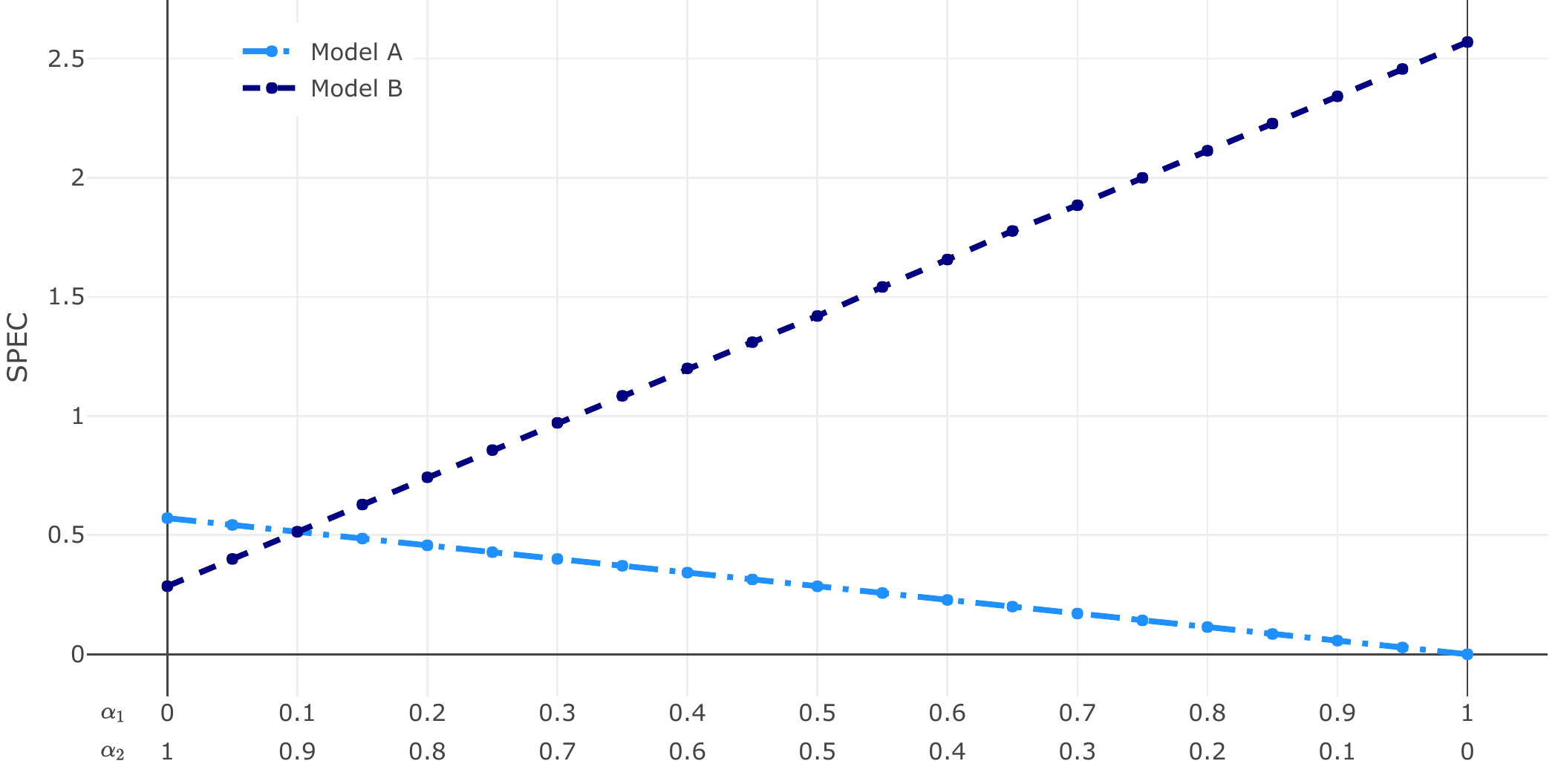}
	% figure caption is below the figure
	\caption{Sensitivity of SPEC values for varying parameters $\alpha_1$ and $\alpha_2$}
	\label{fig: spec_param_sensitivity}       % Give a unique label
\end{figure*}

By adjusting the weights $\alpha_1$ and $\alpha_2$, the sensitivity of SPEC can be adapted to respective costs of a particular application. Maintaining $\alpha_1 = 1-\alpha_2$ resulting SPEC values for the two exemplary models (see Figure~\ref{fig: spec_sample}) are shown in Figure~\ref{fig: spec_param_sensitivity} on page \pageref{fig: spec_param_sensitivity}.
Parameter $\alpha_1$ is used to address the impact of opportunity costs for unfullfilled customer orders, while $\alpha_2$ is adjusting the stock-keeping costs.
The figure emphasizes that higher valuation of stock-keeping costs results in a higher SPEC for model~A and a lower SPEC for model~B in the considered case. This is due to the fact that for model~A stock-keeping costs for eight units occur, whereas for model~B costs arise for four units. Conversely, model~A does not incur any opportunity costs at all, while model~B---as a consequence of underestimating the actual demand at time unit 9---incurs a series of costs associated with the limited satisfiability of customer orders. Hence, the position of the curves of different predictive models relative to each other can be utilized to analyze whether one specific model generally outperforms competing ones or decipher which parameter ratio $\alpha_1 / \alpha_2$ on a particular model performs best.
Therefore, we suggest that the parameters $\alpha_1$ and $\alpha_2$  be chosen depending on the application characteristics and related cost.

\section{Evaluation}
\label{sec: evaluation}

Our proposed novel SPEC metric is evaluated on the basis of simulated and real demand time series data originating from the automotive aftermarket. In particular, we address the reliability and validity of the metric. Additionally, we evaluate the expense, understandability and the relationship to decision making.

\subsection{Data set}

In order to be able to cover as many scenarios of lumpy and intermittent time series and associated forecasts as possible, the time series and their predictions used for evaluating are simulated. Time series from real use cases alone are insufficient in covering all possible scenarios. First, the actual values of a fixed-length time series are simulated, whereas the number of values greater than zero as well as the actual values are normally distributed. In contrast, the arrangement of values greater than zero on the time series of a predefined length follows a uniform distribution.
This ensures that the number of non-zero values, the horizontal as well as the vertical distribution of the simulated time series are random.

In addition, a second part of the evaluation is based on real demand data from spare parts of an automobile manufacturer. The mobility of the spare parts varies greatly depending on product characteristics. This means that both very rarely and more frequently requested products are contained in time series over a gathering period of about two years.

\subsection{Reliability}

Reliability describes a measure of formal accuracy or consistency of scientific measurements. It is part of the variance which can be explained by actual differences in the characteristic that are being measured and not by measurement errors. Highly reliable results must be free of random errors. Thus, the reproducibility of results under the same conditions is guaranteed. However, reliability can not be measured, but only estimated \cite{Carmines1979, Golafshani2003}.

For this reason, there are a number of factors and corresponding approaches to investigate the reliability of a measuring instrument without measuring it directly.

One of these is the repeatability or test-retest reliability, which is determined by repeating the same measurement at a different time \cite{Carmines1979}. Obviously, the repeated application of the proposed SPEC metric always yields the same result, caused by its deterministic character. Thus, this requirement for reliability is met.

Additionally, it must be ensured not only for repeated evaluation on equal time series (actual demand and forecast), but also for similar time series, that similar results are produced. Thus, we simulate a set of different time series representing the actual demand in different scenarios. To do this, we choose appropriate parameters for $\mu$ and $\sigma$ in normal distributions, which control the horizontal and vertical arrangement of demand values. For each of these simulated actual demand series, we simulate several forecasts matching the actual time series by adding a normal distributed error in both the vertical and horizontal directions. Therefore, the parameter $\sigma$ of the two distributions influences the variance of the artificially generated prediction error.

Simulation results show that with a small variance of the artificially induced prediction error, the variance of the SPEC values is low, whereas with increasing variance in error, the variance of the metric also increases. Pearson's correlation coefficient \cite{sedgwick2012pearson} between generated and measured variance in 5,000 simulated time series and 1,000 simulated forecasts for each simulated actual time series with varying error variance in both horizontal and vertical directions yields 0.996, whereas in the same data set other metrics perform comparably well.

In addition, we also evaluated the variances on real data, which are produced by a na\"{i}ve forecast on multiple extracts from different time series of products of different mobility. The intuition behind this assumption is that the characteristics of a time series over randomly selected excerpts remain basically the same, whereas different time series depicting the demand of different parts show structural differences. The results of Levene's test \cite{Levene1960} confirm that the variances of SPEC values assessing different segments of each time series are significantly (p-value~\textless~0.001) below the variance of SPEC values between the different time series. The traditional metrics considered show comparable results. Thus, it can be deduced that the repeated evaluation on different segments of a respective time series leads to only slightly differing SPEC values, which indicates a high reliability.

Thus, it can be confirmed that the SPEC metric achieves reliable results that match those of traditional metrics.

\subsection{Validity}

While the reliability describes that the same results are produced by repeating a measurement, the validity describes whether the correct variable is measured. In addition to reliability, validity is another quality criterion of statistical measurement methods. It generally describes the accuracy to which extent a feature being measured is actually measured. Measurements must be measured accurately to be interpreted in a meaningful way. In this respect, validity is a necessary criterion for assessing the quality of a measure \cite{Carmines1979, Golafshani2003}.

Assessing the validity of a metric requires information about the true object of a measurement, which is difficult to determine. We therefore assume that the SPEC metric should measure the prediction error as precisely as possible with regard to accruing opportunity and stock-keeping costs. Thus, it can be assumed that by overestimating the actual demand, storage costs arise. Similarly, by shifting the prediction---relative to the actual demand---to the left results in premature coverage of demand, which results in stock-keeping costs. However, shifting the prediction to the right results in delayed coverage, which leads to opportunity costs. Based on these assumptions, we simulate another set of random demand time series and associated forecasts. According to the assumptions, the prediction error is varied by adjusting the parameters $\mu$ of the normal distributions which provide either the horizontal or vertical shift of the prediction error.

For the sole change of the vertical error by adaptation of the parameter $\mu$, an average correlation coefficient between the values of $\mu$ and all metrics considered (including SPEC) is 0.999. This results from 5,000 simulated actual time series and 1,000 simulated predictions each with constant variance of the prediction error. This result shows that all considered metrics map a vertical increasing under- or overestimation equally well.

In contrast, changing the horizontal error by adjusting the parameter $\mu$, which provides for shifting the prediction relative to the demand times, shows a different picture. The parameter $\mu$ and the values of the traditional metrics are either uncorrelated (MAE, RMSE, MASE) or not calculable at all (MAPE, sMAPE). Thus, the SPEC metric performs much better yielding a correlation coefficient of 0.867. Hence, the resulting stock-keeping and opportunity costs are addressed significantly better.

In a second step, we utilize real data from the automotive aftermarket to make sure the proposed metric actually measures what it should measure. To do this, we calculate the na\"{\i}ve forecast of randomly selected extracts of actual demands and calculate the actual costs based on empirical values for costs caused by inaccuate predictions. It becomes clear that the SPEC metric ($r$~=~0.964, p-value~\textless~0.001), in contrast to traditional metrics (e.g., MAE: $r$~=~-0.021, p-value~\textgreater~0.05), represents much better the true costs to be assumed.

\subsection{Additional criteria}

In addition to the criteria of validity and reliability---both essential for a metric---we discuss a few rather soft criteria.

The significantly more complex calculation of the SPEC metric compared to the traditional metrics results in a certain disadvantage with regard to the calculation performance. Nevertheless, all of these metrics can be calculated on today's computers without noticeable performance differences.

Similarly, some of the traditional metrics have an advantage over SPEC in terms of understandability. Notably the MAE which is very simple compared to the SPEC which appears more complex especially after considering horizontal deviations. Nevertheless, the aspect that SPEC represents stock-keeping and opportunity costs is conducive to understandability.

An absolute advantage of SPEC over traditional metrics is the direct link to business aspects and decision-making. Since cost equivalents can be addressed by the choice of the parameters $\alpha_1$ and $\alpha_2$, the costs accrued by a prediction error can be interpreted directly. This peculiarity makes SPEC the ideal metric when it comes to evaluating demand forecasts where prediction errors cause stock-keeping and/or opportunity costs.

\section{Robustness Check}
\label{sec: robustness check}

In order to demonstrate the robustness of our proposed SPEC metric we evaluate it on an additional real-world data set that did not apply during the development of the presented metric. In addition to the simulated data sets and the automotive spare parts case already elaborated on in the previous section, this data set comes from five different supermarket branches in the food retail sector and describes the hourly demand for a specific, rather sporadically required product group over one year \cite{walk2020}. The data set shows a total of about 25\% zero values, which do not follow any recognizable regularity across all branches. In addition, we have approximated numbers about actual storage costs of the respective product group and corresponding opportunity costs represented by profit losses. As in the automotive spare parts case considered in Section \ref{sec: evaluation}, a ratio of 3 to 1 between opportunity costs and stock-keeping costs seems to be an appropriate choice for the parameters $\alpha_1$ and $\alpha_2$.

Analogous to the procedure in Section \ref{sec: evaluation}, we point out that reliability and validity are also given on this data set. The results of Levene's test show that the variance of the prediction error measured by SPEC over different segments of one time series are significantly (p-value~\textless~0.001) smaller than the variance over different time series from different branches. This means that, assuming consistent characteristic demand patterns within a branch, the reliability of the proposed metric can also be confirmed on this data set. In addition, SPEC also rates the actual costs significantly better than traditional metrics do ($SPEC_{0.75,0.25}$: $r$~=~0.912, p-value~\textless~0.001; e.g., MAE: $r$~=~0.201, p-value~\textgreater~0.05).

\section{Conclusion}
\label{sec: conclusion}
The most accurate forecast of product demand is essential for the optimization of logistics and production. In order to optimally train predictive models, which account for this requirement, the forecast compared to the actual demand needs to be assessed by a proper metric. However, if a metric does not represent the actual prediction error, predictive models are insufficiently optimized and, consequently, will yield inaccurate predictions. To overcome this issue, we have developed a novel accuracy measure called Stock-keeping-oriented Prediction Error Costs (SPEC).

This metric takes into account not only deviations of the forecast relative to the actual demand in vertical direction but also in horizontal direction. Thus, the SPEC metric addresses the costs incurred by a company in the case of early or late delivery to the customer, resulting either in stock-keeping costs or opportunity costs. Especially for these types of deviations, the results of the evaluation show that the metric achieves significantly better results in terms of validity than traditional metrics. Due to the generic formulation any cost equivalent is allowed to be mapped by choosing proper $\alpha$ parameters. The fields of applications of the SPEC metrics are therefore not limited to the automotive aftermarket. Additionally, it is applicable to all possible areas where there is a partial sporadic demand. Examples include certain products in supermarkets, online retail, gastronomy or even services for which either costs for late or too early delivery, caused by a forecast error, may be incurred.

Nevertheless, our proposed metric also has limitations. To critically summarize, SPEC considers fictive opportunity and stock-keeping costs. Especially in time series scenarios with a forecast horizon of $n > 1$, this error metric is a significantly more representative accuracy measure compared to others. One can argue, that SPEC is penalizing missing parts until the end of the forecast horizon even though in reality, for example, SKUs with high urgency can be delivered via express orders. However, costs arise for such express orders as well and can be outlined by SPEC.

Overall, we are convinced that the proposed metric is a viable and appropriate alternative to traditional accuracy measures. We therefore recommend readers from academia as well as practitioners to approach the SPEC metric in future studies as an additional prediction error measure for lumpy and intermittent demand forecasts.

% if added before the last page, this command can help balancing columns
%\addtolength{\textheight}{-.2cm} 

%Bibliography 
\bibliographystyle{ieeetr}
\bibliography{references}

\end{document}

%% file: hicss51-packages.tex
\usepackage[letterpaper]{geometry}
\usepackage{hicss51}
\usepackage{times}
\usepackage[none]{hyphenat}
\usepackage{url}
\usepackage{latexsym}
\usepackage{indentfirst}
\usepackage{graphicx}
\graphicspath{{images/}}

%% file: hicss51.bbl
\begin{thebibliography}{10}

\bibitem{Mahmoud1984}
E.~Mahmoud, ``{Accuracy in forecasting: A survey},'' {\em Journal of
  Forecasting}, vol.~3, no.~2, pp.~139--159, 1984.

\bibitem{Kim2016}
S.~Kim and H.~Kim, ``{A new metric of absolute percentage error for
  intermittent demand forecasts},'' {\em International Journal of Forecasting},
  vol.~32, no.~3, pp.~669--679, 2016.

\bibitem{Wallstrom2010}
P.~Wallstr{\"{o}}m and A.~Segerstedt, ``{Evaluation of forecasting error
  measurements and techniques for intermittent demand},'' {\em International
  Journal of Production Economics}, vol.~128, no.~2, pp.~625--636, 2010.

\bibitem{Waller2016}
D.~Waller, ``{Methods for Intermittent Demand Forecasting},'' pp.~1--6, 2016.

\bibitem{Kourentzes2013}
N.~Kourentzes, ``{Intermittent demand forecasts with neural networks},'' {\em
  International Journal of Production Economics}, vol.~143, no.~1,
  pp.~198--206, 2013.

\bibitem{Mukhopadhyay2012}
S.~Mukhopadhyay, A.~O. Solis, and R.~S. Gutierrez, ``{The accuracy of
  non-traditional versus traditional methods of forecasting lumpy demand},''
  {\em Journal of Forecasting}, vol.~31, no.~8, pp.~721--735, 2012.

\bibitem{Vaishnavi2004}
V.~K. Vaishnavi and W.~L. Kuechler, ``{Design Science Research in Information
  Systems},'' {\em Ais}, pp.~1--45, 2004.

\bibitem{yan2012toward}
W.~Yan, ``{Toward automatic time-series forecasting using neural networks},''
  {\em IEEE Transactions on Neural Networks and Learning Systems}, vol.~23,
  no.~7, pp.~1028--1039, 2012.

\bibitem{makridakis1982accuracy}
S.~Makridakis, A.~Andersen, R.~Carbone, R.~Fildes, M.~Hibon, R.~Lewandowski,
  J.~Newton, E.~Parzen, and R.~Winkler, ``{The accuracy of extrapolation (time
  series) methods: Results of a forecasting competition},'' {\em Journal of
  forecasting}, vol.~1, no.~2, pp.~111--153, 1982.

\bibitem{WERON2014}
R.~Weron, ``{Electricity price forecasting: A review of the state-of-the-art
  with a look into the future},'' {\em International Journal of Forecasting},
  vol.~30, no.~4, pp.~1030--1081, 2014.

\bibitem{HONG2014357}
T.~Hong, P.~Pinson, and S.~Fan, ``{Global Energy Forecasting Competition
  2012},'' {\em International Journal of Forecasting}, vol.~30, no.~2,
  pp.~357--363, 2014.

\bibitem{kalekar2004time}
P.~S. Kalekar, ``{Time series forecasting using holt-winters exponential
  smoothing},'' {\em Kanwal Rekhi School of Information Technology},
  vol.~4329008, no.~13, 2004.

\bibitem{Chen2004}
Z.~Chen and Y.~Yang, ``{Assessing Forecast Accuracy Measures},'' 2004.

\bibitem{Armstrong1992}
J.~Armstrong and F.~Collopy, ``{Error measures for generalizing about
  forecasting methods: Empirical comparisons},'' {\em International Journal of
  Forecasting}, vol.~8, no.~1, pp.~69--80, 1992.

\bibitem{Shcherbakov2013}
M.~V. Shcherbakov, A.~Brebels, N.~L. Shcherbakova, A.~P. Tyukov, T.~A.
  Janovsky, and V.~A.~e. Kamaev, ``{A survey of forecast error measures},''
  {\em World Applied Sciences Journal}, vol.~24, no.~24, pp.~171--176, 2013.

\bibitem{Hyndman2006}
R.~J. Hyndman and A.~B. Koehler, ``{Another look at measures of forecast
  accuracy},'' {\em International Journal of Forecasting}, vol.~22, no.~4,
  pp.~679--688, 2006.

\bibitem{MAKRIDAKIS1993}
S.~Makridakis, ``{Accuracy measures: theoretical and practical concerns},''
  {\em International Journal of Forecasting}, vol.~9, no.~4, pp.~527--529,
  1993.

\bibitem{Goodwin1999}
P.~Goodwin and R.~Lawton, ``{On the asymmetry of the symmetric MAPE},'' {\em
  International Journal of Forecasting}, vol.~15, no.~4, pp.~405--408, 1999.

\bibitem{Carmines1979}
E.~G. Carmines and R.~A. Zeller, {\em {Reliability and Validity Assessment}},
  vol.~17.
\newblock Sage publications, 1979.

\bibitem{Golafshani2003}
N.~Golafshani, ``{Understanding Reliability and Validity in Qualitative
  Research},'' {\em The Qualitative Report}, vol.~8, no.~4, pp.~597--606, 2003.

\bibitem{sedgwick2012pearson}
P.~Sedgwick, ``Pearson’s correlation coefficient,'' {\em Bmj}, vol.~345,
  p.~e4483, 2012.

\bibitem{Levene1960}
H.~Levene, ``{Robust Tests for Equality of Variances},'' in {\em Contributions
  to Probability and Statistics: Essays in Honor of Harold Hotelling}
  (I.~Olkin, ed.), pp.~278--292, Palo Alto: Stanford University Press, 1960.

\bibitem{walk2020}
J.~Walk, R.~Hirt, N.~K{\"u}hl, and E.~R. Hersl{\o}v, ``Half-empty or half-full?
  a hybrid approach to predict recycling behavior of consumers to increase
  reverse vending machine uptime,'' in {\em Exploring Service Science}
  (H.~N{\'o}voa, M.~Dr{\u{a}}goicea, and N.~K{\"u}hl, eds.), (Cham),
  pp.~107--120, Springer International Publishing, 2020.

\end{thebibliography}
